\documentclass[journal, twoside, web]{ieeecolor}
\usepackage{generic}
\usepackage{cite}
\usepackage{amsmath,amssymb,amsfonts}
\usepackage{algorithmic}
\usepackage{graphicx}
\usepackage{algorithm,algorithmic}
\usepackage{hyperref}
\hypersetup{hidelinks=true}
\usepackage{textcomp}
\usepackage{caption}
\usepackage{subcaption}
\usepackage{comment}
\newcommand{\figref}[1]{Fig.~\ref{fig:#1}}
\newcommand{\secref}[1]{Section~\ref{sec:#1}}
\newcommand{\tblref}[1]{Table~\ref{tbl:#1}}

\newcommand{\eqnref}[1]{Equation~\ref{eq:#1}}

\def\BibTeX{{\rm B\kern-.05em{\sc i\kern-.025em b}\kern-.08em
    T\kern-.1667em\lower.7ex\hbox{E}\kern-.125emX}}
\begin{document}
\title{Multimodal Physical Activity Forecasting in Free-Living Clinical Settings: Hunting Opportunities for Just-in-Time Interventions\\}
\author{Abdullah Mamun$^{1,2}$, Krista S. Leonard$^{1}$, Megan E. Petrov$^{3}$, Matthew P. Buman$^{1}$, Hassan Ghasemzadeh$^{1}$%
\thanks{Authors are with Arizona State University, Phoenix, AZ 85054, USA
        {email: \{a.mamun, krista.leonard, megan.petrov, mbuman,\\ hassan.ghasemzadeh\}@asu.edu}}
\thanks{$^{1}$College of Health Solutions, Arizona State University}
\thanks{$^{2}$School of Computing and Augmented Intelligence, Arizona State University}
\thanks{$^{3}$Edson College of Nursing and Health Innovation, Arizona State University}%
}
\maketitle
\begin{abstract}
\textit{Objective}: This research aims to develop a lifestyle intervention system, called \textit{MoveSense}, that forecasts a patient's activity behavior to allow for early and personalized interventions in real-world clinical environments.
\textit{Methods}: We conducted two clinical studies involving 58 prediabetic veterans and 60 patients with obstructive sleep apnea to gather multimodal behavioral data using wearable devices. We develop multimodal long short-term memory (LSTM) network models, which are capable of forecasting the number of step counts of a patient up to 24 hours in advance by examining data from activity and engagement modalities. Furthermore, we design goal-based forecasting models to predict whether a person's next-day steps will be over a certain threshold. \textit{Results}: Multimodal LSTM with early fusion achieves 33\% and 37\% lower mean absolute errors than linear regression and ARIMA respectively on the prediabetes dataset. LSTM also outperforms linear regression and ARIMA with a margin of 13\% and 32\% on the sleep dataset. Multimodal forecasting models also perform with 72\% and 79\% accuracy on the prediabetes dataset and sleep dataset respectively on goal-based forecasting.
\textit{Conclusion}: Our experiments conclude that multimodal LSTM models with early fusion are better than multimodal LSTM with late fusion and unimodal LSTM models and also than ARIMA and linear regression models.
\textit{Significance}: We address an important and challenging task of time-series forecasting in uncontrolled environments. Effective forecasting of a person's physical activity can aid in designing adaptive behavioral interventions to keep the user engaged and adherent to a prescribed routine.
\end{abstract}


\section{Introduction}
\label{sec:introduction}
Physical activity is a crucial factor that can reduce the risk of many chronic conditions such as cardiovascular diseases, diabetes, and cancers \cite{atienza2011self, haskell2007physical}. However, it remains challenging for many individuals to maintain a consistent routine of an active lifestyle \cite{seefeldt2002factors}. As a result, one-third of people aged 15 years and older lack sufficient physical activities \cite{park2020sedentary}. Among many different ways of promoting physical activity, recently mHealth interventions have become a popular choice \cite{gal2018effect, romeo2019can, schoeppe2016efficacy}. Previous research suggests that engagement with lifestyle intervention applications is a strong contributor to the success of these interventions. Usage of a particular intervention application can be associated with healthier behaviors such as healthy eating and increased physical activity \cite{gilliland2015using, wang2015wearable, kirwan2012using}. It is believed and also supported by evidence that activity prompts and engagement reminders can improve adherence to physical activity \cite{wolner2021can}. However, designing and automating effective interventions that are also tailored to individual needs requires knowing in advance when a user is not likely to follow the recommended level of physical activity on a particular day. 

This paper investigates how accurately it is possible to forecast a person's physical activity, thus enabling adaptive and personalized behavioral interventions. In particular, we study both unimodal time-series models based on physical activity and multimodal models based on physical activity and engagement in mHealth interventions in our activity forecasting design. The engagement metrics are computed based on the amount of involvement in intervention apps. We hypothesize that multimodal models carry richer behavioral data and therefore offer an improved forecasting performance compared to unimodal models. To the best of our knowledge, our work is the first that studies time-series multimodal activity forecasting in the context of mHealth lifestyle interventions \cite{mamun2022multimodal}. To address this gap in research, we formulate a time-series forecasting problem and provide a formal definition of the problem. We propose a behavioral intervention system, MoveSense\footnote{Code available: \textit{https://github.com/ab9mamun/MoveSense}}, that utilizes LSTM (long short-term memory network) based neural networks to forecast the daily number of steps of a person one day in advance.

Our extensive analyses using data collected in two clinical studies show that an LSTM model can forecast the number of steps a user will walk the next day with a mean absolute error of as low as 1677, as shown in \secref{most_engaged}. The details of the dataset and the experimental procedures have been discussed in \secref{methods}. At a glance, our contributions in this article include (1) formulation of a time-series forecasting problem for activity forecasting in behavioral health; (2) implementing the MoveSense system with LSTM-based forecasting solutions and testing them on two datasets; (3) a comparison among forecasting with different modalities and different fusion methods, i.e. how the features of different modalities or their representations are combined; (4) providing analytical results demonstrating that engagement and physical activity metrics are correlated; and (5) making our clinical data and software tools available for public use upon publication.

\section{Related Work}
\subsection{Time-Series Forecasting}
Time-series forecasting is an age-old problem that has been solved in many different contexts in recent times and the past few decades. Traditional models, such as nearest neighbor algorithms, autoregressive models, and the autoregressive integrated moving average (ARIMA) method were some of the popular choices before the popularity of deep neural networks \cite{de200625,fernandez1999exchange,zhang2003time}.  A regression tree-based method \cite{minor2017forecasting} proved to be able to predict the time when a particular event will take place in a smart home setting. However, in recent times, more and more forecasting research has been driven by deep learning methods, especially recurrent neural networks (RNN) \cite{madan2018predicting, smyl2020hybrid, zeroual2020deep, mamun2022multimodal} and transformers \cite{woo2024unified, zeng2023transformers}. Among the RNN-based models, gated recurrent unit (GRU) and long short-term memory (LSTM) networks are most commonly used. A comparative study \cite{siami2018comparison} tested LSTM and ARIMA models on several stock market datasets and showed that LSTM  beats ARIMA by a large margin. LSTM and transformers have been used in many different time-series forecasting and prediction problems including energy consumption \cite{dubey2021study}, traffic \cite{lai2018modeling}, sales and demand \cite{qi2021known}, weather and state of electrical machinery \cite{zhou2021informer}, hard drive failure \cite{dos2017predicting}, etc.

Although mostly popular for image and video processing tasks, convolutional neural networks (CNN) can be used to solve problems with time-series data as well. For example, LSTNet \cite{lai2018modeling} is a deep neural network that combines CNN, LSTM, and autoregressive layers to forecast traffic, electricity consumption, solar energy, and exchange rate. CNN can also be used for the imputation of missing sensor data \cite{mamun2022designing}. MTEX-CNN \cite{assaf2019mtex} is a CNN-based forecasting model that also provides explanations. CNNs have been used for many other time-series analysis and forecasting problems such as website visits \cite{wibawa2022time} and analyzing financial data \cite{chen2016financial}. Despite the reasonably good performance of these methods, they are not free from limitations. Different methods have been used for forecasting hospital admission \cite{zhou2018time}, mortality \cite{deng2022explainable}, childbirth risks \cite{mamun2023neonatal}, etc. However, forecasting for clinical applications has not advanced in the same way as other domains of forecasting with machine learning. Furthermore, to the best of our knowledge, activity forecasting using wearable sensor data and smartphone app usage has not been investigated enough to explore its promises \cite{mamun2022multimodal}.

\subsection{Physical Activity Monitoring}
The intensity and duration of physical activity can be positively affected through self-monitoring and logging. \cite{wang2015wearable} found that participants who self-reported using their Fitbit tracker and app frequently also had greater increases in their physical activity. In a user study, it was observed that participants who used the provided application to record daily steps every day were more likely to log at least 10,000 steps compared to those who used the app less \cite{kirwan2012using}. However, these studies are limited by the use of self-reported measures. Wang et al. included a self-reported measure of app usage where participants would report how often they used the app \cite{wang2015wearable}. \cite{kirwan2012using} included a self-reported measure of step counts suggesting that their findings are more focused on how app usage was associated with logging in steps rather than actually obtaining a certain step count. In contrast to these studies, \cite{edney2019user} used system usage data to objectively measure app usage, i.e., the number of times the app features were used. They found that interactive apps (e.g., gamification) are more likely to increase user engagement and app usage and that increased usage is associated with greater increases in physical activity over time.

BeWell24 is a multicomponent smartphone app intervention that targets behavior change in the 24h-spectrum (i.e., sleep, sedentary time, physical activity) \cite{buman2016bewell24}. It was developed using a user-centered iterative design framework that included quantitative process-level outcomes related to app usage and post-intervention qualitative feedback from users with respect to app design and satisfaction. The study found that overall, satisfaction with the app was modest. Given the promising metrics of app usage of the BeWell24, the next logical step is to examine the association of app use with intervention outcomes such as physical activity. SleepWell24 is a similar smartphone app that was developed to encourage its users to monitor their continuous positive airway pressure (CPAP) usage and sleep quality in addition to their physical activity levels \cite{petrov2020rationale}.

\subsection{Engagement with Technology}
The definition of engagement with apps can have one of the many metrics including the frequency of use, duration of use, number of log-ins, and pages viewed \cite{short2018measuring, perski2017conceptualising}. Unfortunately, although the data are collected, many studies report limited or no app usage metrics. Consequently, researchers have made a call for future studies to report app usage metrics \cite{schoeppe2016efficacy}. Therefore, this study aims to expand upon the limited literature and examine app usage metrics of a multicomponent smartphone app intervention targeting lifestyle behaviors across 24 hours and its association with physical activity. We hypothesize that a person's physical activity is correlated with when, how, and how often they use a fitness tracker app.

\subsection{Multimodal Learning}
Multimodal learning models have the potential to make predictions with more confidence than unimodal models. When there is more than one modality of input, a multimodal learning architecture is necessary to process them effectively. However, multimodal learning comes with five different challenges that need to be addressed: representation, translation, alignment, fusion, and co-learning \cite{baltruvsaitis2018multimodal}. We refer to \cite{baltruvsaitis2018multimodal} for details of those challenges. Previously, it was found that a multimodal speech classifier that uses both audio and video can classify a speech with higher accuracy than a similar unimodal classifier that uses only audio or only video input \cite{ngiam2011multimodal}.  In multimodal learning, features from different modalities or their encodings need to be combined at some point, which is known as fusion. Fusion can be done in many different ways, e.g. early fusion, late fusion, mid-level fusion, etc. \cite{kline2022multimodal} \cite{roitberg2019analysis}. For a multimodal gesture recognition problem from video and depth data, it has been found that the test accuracy can be higher in a cross-stitch fusion technique than in late fusion and early or mid-level fusion at different layers \cite{roitberg2019analysis}.

\section{Methods}
\label{sec:methods}
\subsection{MoveSense System Overview}
\begin{figure}[!t]
\centerline{\includegraphics[width=0.99\columnwidth]{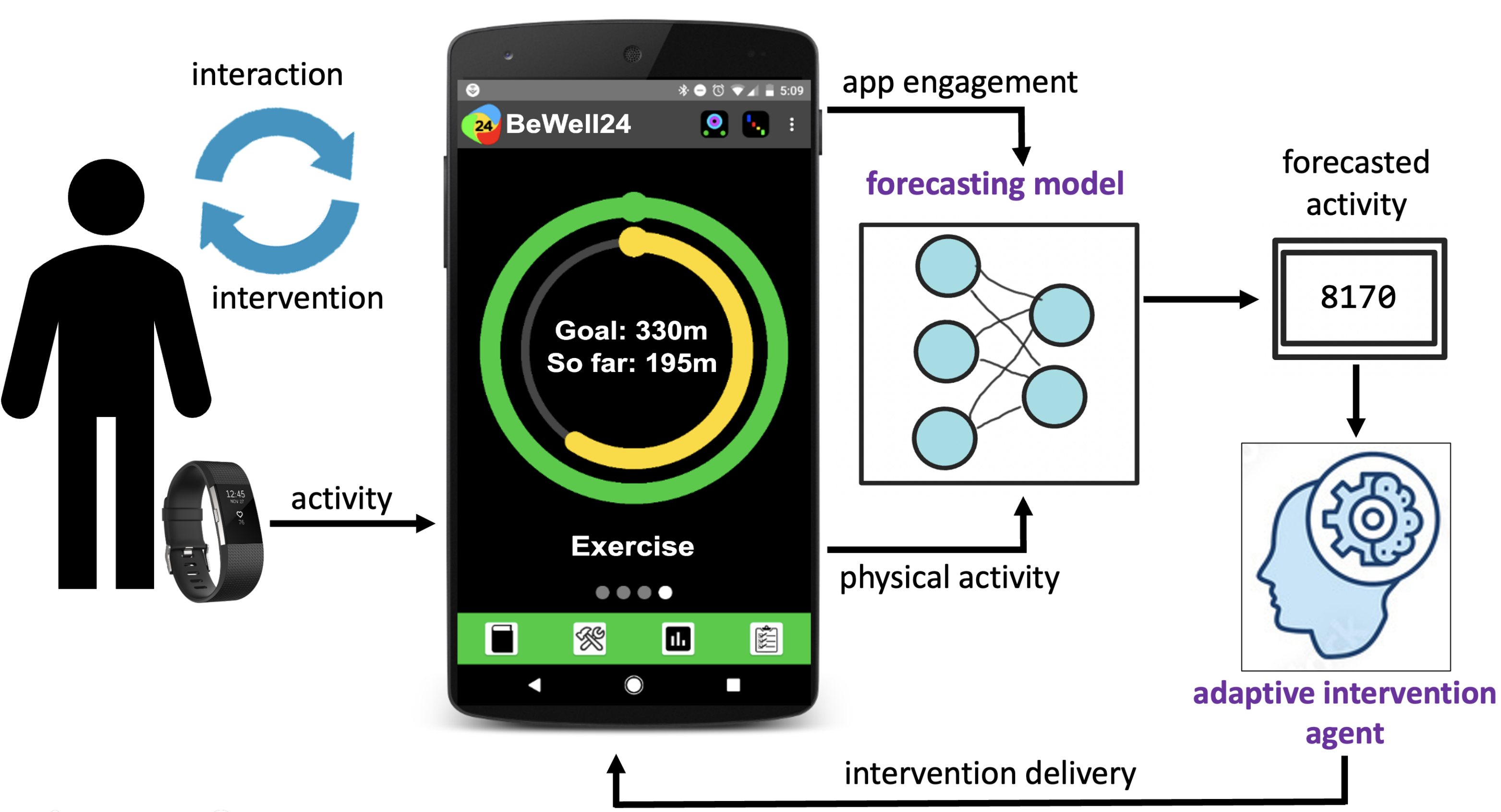}}
\caption{The overview of our MoveSense system and its components: a wearable wristband, a lifestyle intervention app, a forecasting model, and an adaptive intervention agent.}
\label{fig:system}
\end{figure}

MoveSense is a closed-loop and adaptive system that guides its users to a better lifestyle with the help of its core components, an intervention app, an activity tracker, machine learning models for forecasting activities, and an adaptive intervention agent, as shown in \figref{system}. These components are going to work together to facilitate a user to better manage their lifestyle.

\subsubsection{Intervention Applications}
\begin{figure*}[h!]
     \centering
     \begin{subfigure}[b]{0.168\textwidth}
         \centering
         \includegraphics[width=\textwidth]{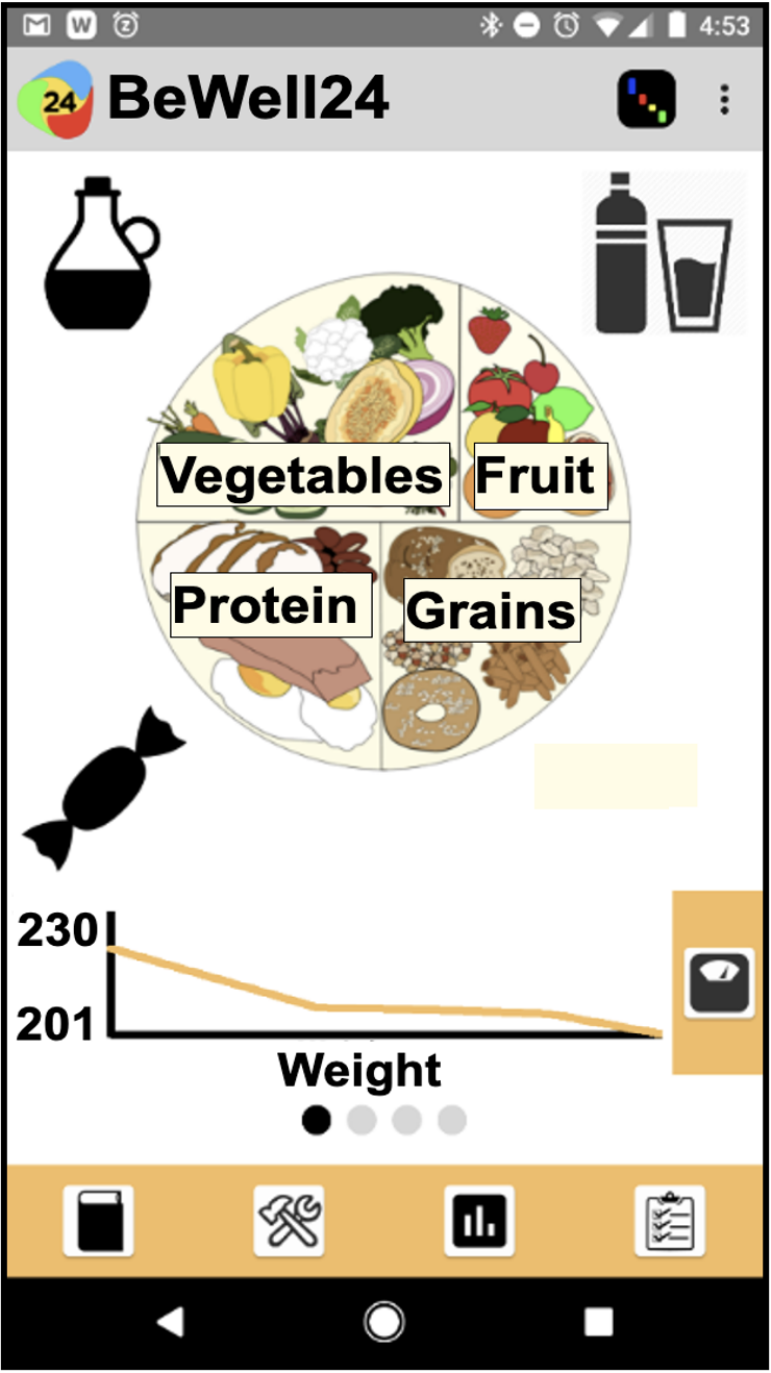}
         \caption{Nutrition}

     \end{subfigure}
     \begin{subfigure}[b]{0.17\textwidth}
         \centering
         \includegraphics[width=\textwidth]{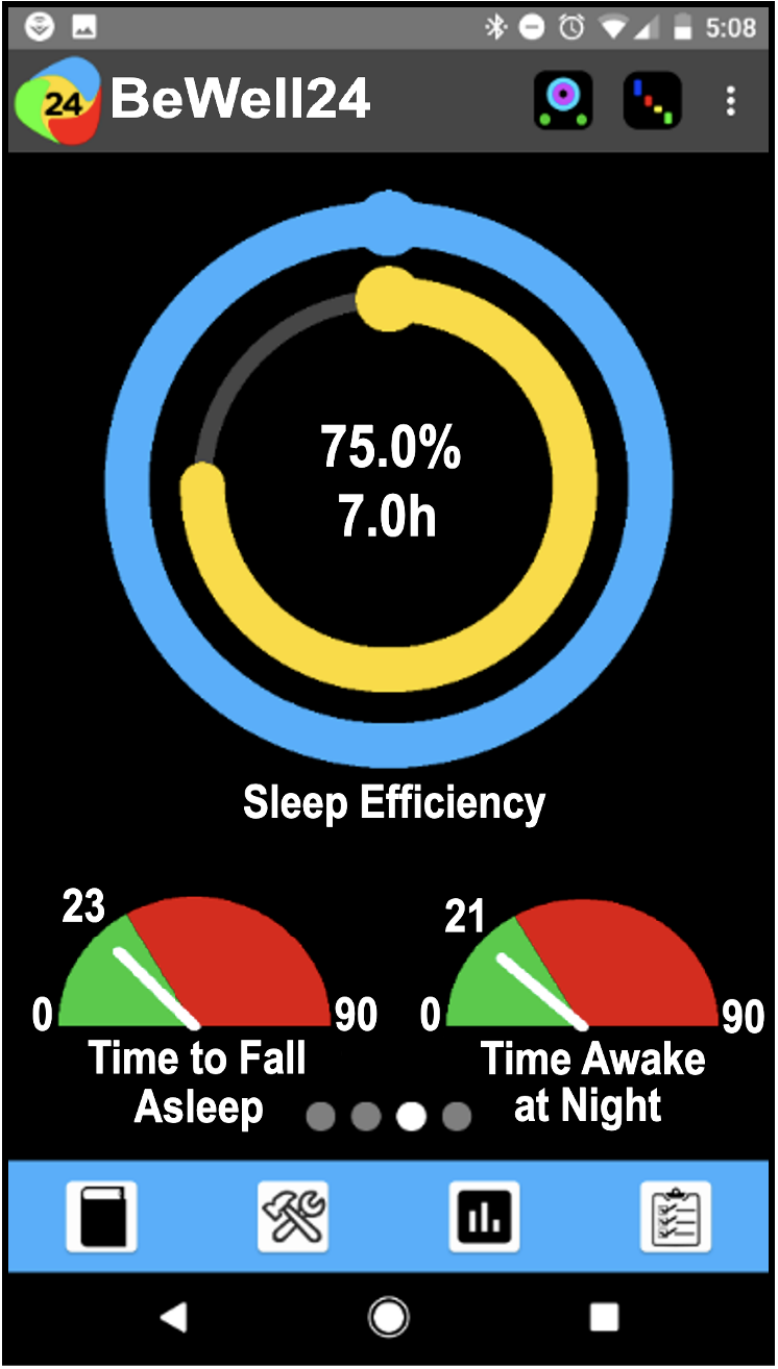}
         \caption{Sleep}
     \end{subfigure}
     \begin{subfigure}[b]{0.17\textwidth}
         \centering
         \includegraphics[width=\textwidth]{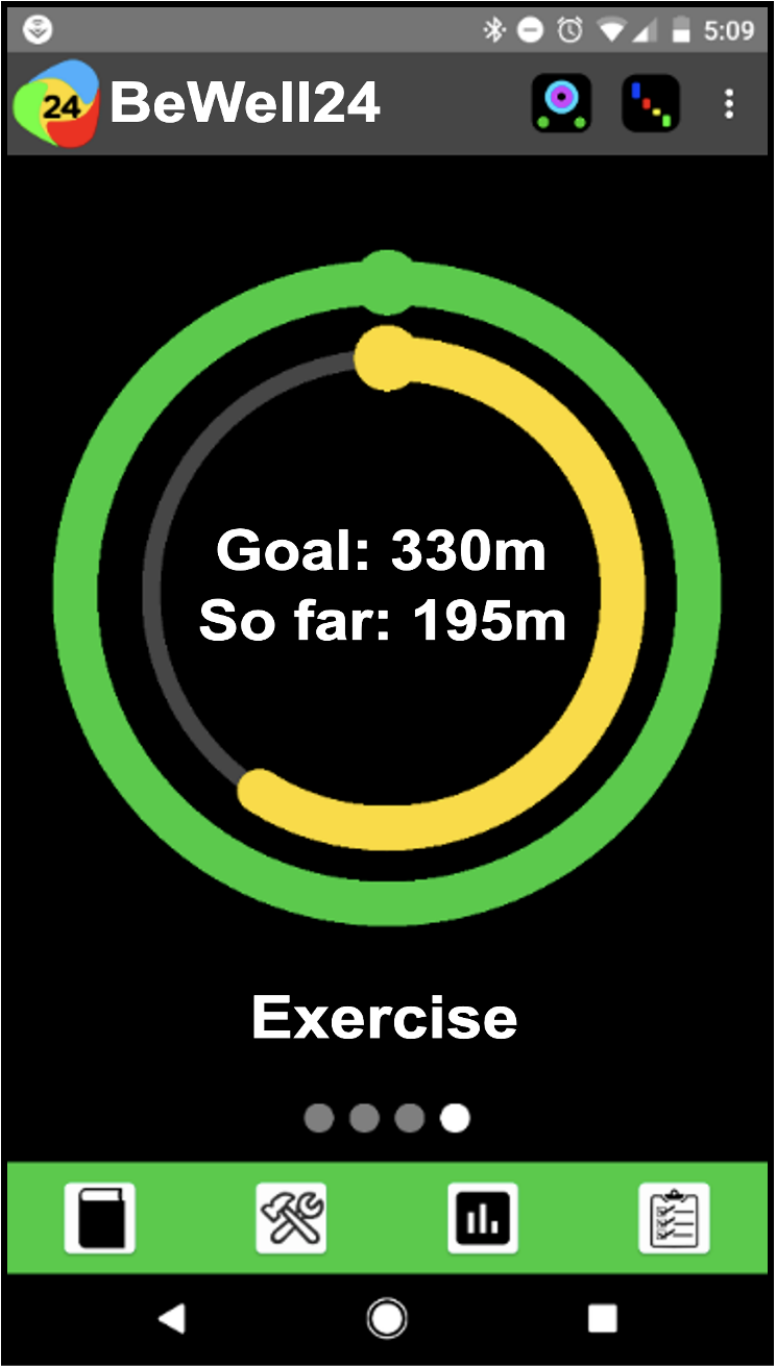}
         \caption{Exercise}

     \end{subfigure}
    \begin{subfigure}[b]{0.1713\textwidth}
        \centering
    \includegraphics[width=\textwidth]{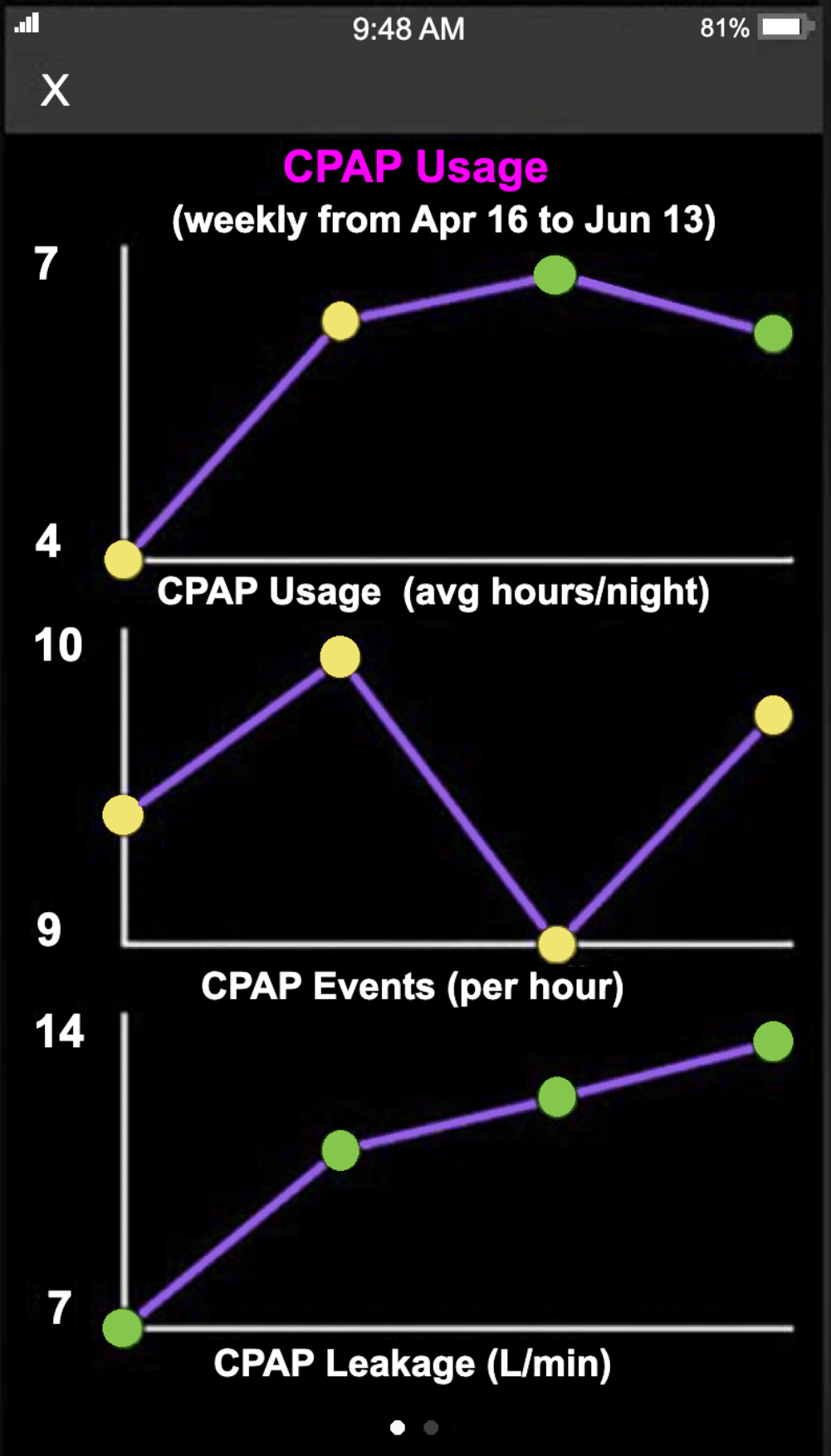}
        \caption{CPAP usage}
        \label{fig:cpap}
     \end{subfigure}     

\caption{Screenshots from the BeWell24 and SleepWell24 intervention applications of our MoveSense system. The BeWell24 app has the features of monitoring physical activity, diet, etc. The SleepWell24 app has an additional component for monitoring the usage of continuous positive airway pressure (CPAP) devices.}
\label{fig:screenshots}
\end{figure*}

A lifestyle intervention app is commonly implemented to assist individuals with achieving their physical activity goals. For example, BeWell24 is a multicomponent smartphone app that was designed to target lifestyle behaviors across 24 hours (i.e., physical activity, sedentary behavior, sleep) to improve glucose metabolism in patients prediabetic for type 2 diabetes. Briefly, the smartphone app included behavioral strategies, based upon evidence-based behavior change techniques (e.g., goal setting, stimulus control, self-regulatory strategies, etc.), for sleep, sedentary time, physical activity, and dietary intake. A user of the app would be able to self-monitor their sleep and physical activity based on objective assessments. They would also receive personal feedback on their behaviors based on self-monitoring. A few screenshots of the application are presented in \figref{screenshots}.

Another smartphone app called SleepWell24 was developed that would allow a user to track their daily duration of physical activity and its breakdown in the durations of sedentary, light, and moderate to vigorous physical activities as seen in \figref{cpap}. Additionally, the app would also  allow a user to monitor their sleep duration as well as the quality of sleep in terms of sleep efficiency and also the number of hours they are going to use the provided CPAP device. The purpose of the app was to improve PAP adherence in people with obstructive sleep apnea with mHealth technologies. The major components of the app are self-monitoring, PAP adherence, sleep, physical activity, diet, and patient-provider communication which enabled easy communication between the patient and their clinical provider \cite{petrov2020rationale}.

\subsubsection{Fitbit Wearable Wristband}
To allow the smartphone apps to monitor the daily physical and sleeping activities of a user, the decision was taken to use Fitbit Charge 2 for continuous objective monitoring of sleep and physical activity. The user would download the Fitbit app on their smartphone and they would be able to view their Fitbit data on the smartphone app. The Fitbit app would be used to sync Fitbit with the user's smartphone and then auto-populate its data into the corresponding smartphone.

\subsubsection{Activity Forecasting}
\label{sec:problemformulation}
At this point, we know what type of data will be available to us and now can design our forecasting model. We consider a supervised learning setting where a machine learning model uses multimodal forecasting to predict the value of a variable at a future time step using different variables from different modalities of the past. For a particular time step $t$, we have different modalities of input, $u_t \in \mathbb{R}^p, v_t \in  \mathbb{R}^q$. Here, $p, q \in \mathbb{N}$ are the dimensions of $u_t$ and $v_t$ vectors, respectively.  We can write, the input at time step $t$, $x_t  = (u_t, v_t)$. The goal is to forecast the output variable for time step $t+1$, $y_{t+1}$   using $w$ number of past input examples, $(x_{t-w+1}, x_{t-w+2}, …, x_t)$. Here, $w$ is the window size. If $f$ is a function that operates the calculation on the input features and predicts the output, we can write,
\begin{equation}
\label{eq:forecast1}
    \hat{y}_{t+1} = f (x_{t-w+1}, x_{t-w+2}, …, x_t)
\end{equation}

In our experiments, $u_t$ represents app engagement features and $v_t$ represents physical activity features. To estimate the function, $f$, we consider the early fusion  and late fusion multimodal learning methods. Early fusion combines the $u_t$ and $v_t$ at the feature level before passing through any layers of a neural network, whereas, the late fusion method processes $u_t$ and $v_t$ separately through different channels of the neural network, and a little before the final prediction, the two channels are combined.

If we unroll \eqnref{forecast1} for different modalities, the formulation will depend on which fusion method is to be used. For shorthand, let $u = (u_{t-w+1}, u_{t-w+2}, …, u_t )$, and $v = (v_{t-w+1}, v_{t-w+2}, …, v_t)$. Suppose, $g(\alpha,\beta)$ is a function that takes two variables $\alpha$ and $\beta$ and fuses them together to create a single variable.  For early fusion, we can write,

\begin{equation}
\label{eq:forecastearly}
    \hat{y}_{t+1} = f_{early} (g(u, v)) 
\end{equation}

This means we will first concatenate u and v before feeding to our forecaster $f_{early}$.
And for late fusion, we can write,

\begin{equation}
    \label{eq:forecastlate}
\hat{y}_{t+1} = f_{late}(g(f_1(u), f_2(v)))	 
\end{equation}

\begin{figure}[!t]
\centerline{\includegraphics[width=0.9\columnwidth]{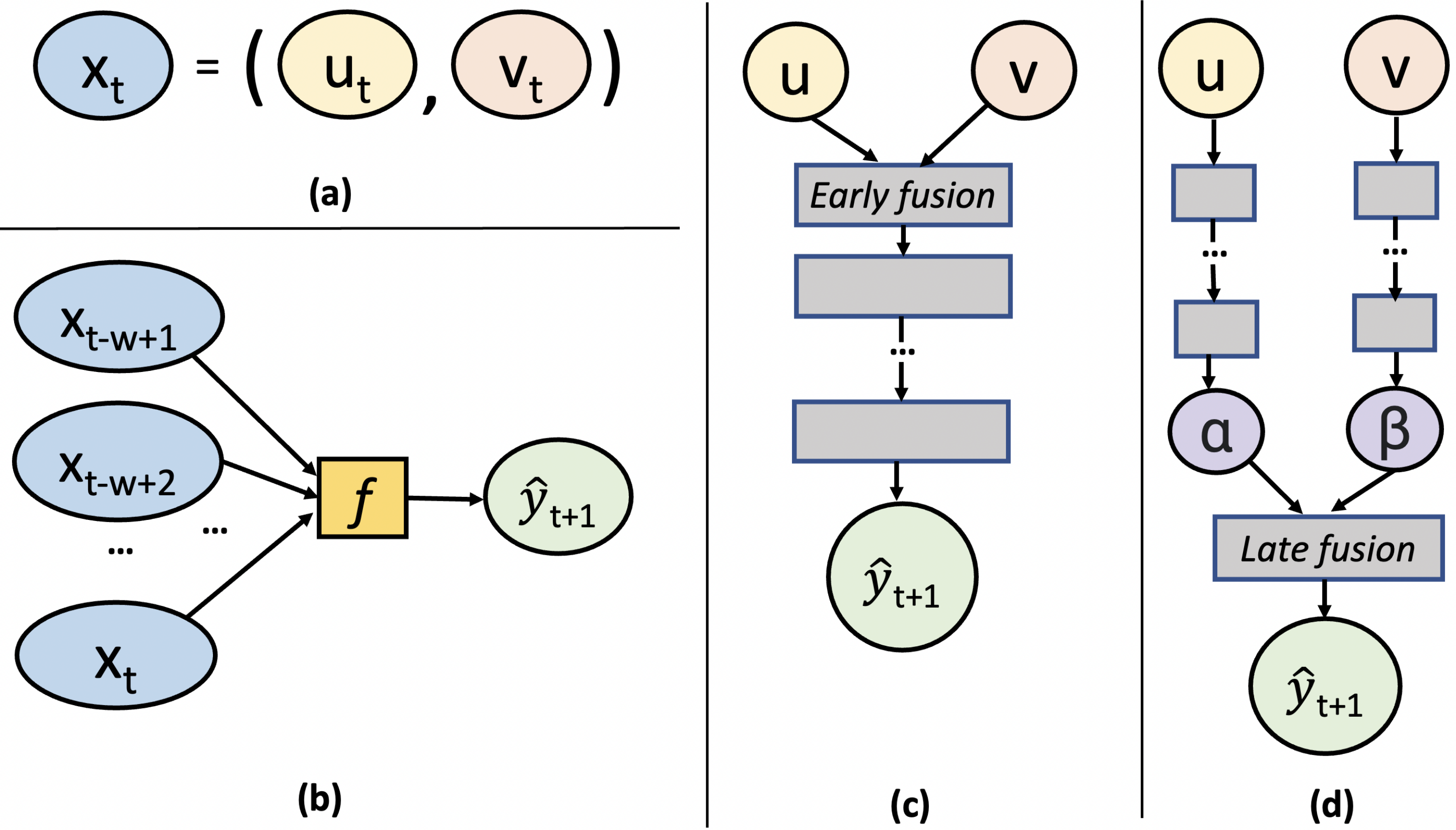}}
\caption{A generic formulation of the multimodal time-series forecasting problem of the MoveSense system. (a) A multimodal feature $x_t$ at time step $t$ is the collection of two features from two modalities, $u_t$ and $v_t$. (b) The forecasting function $f$ takes the values of the last $w$ time steps for the multimodal feature $x$ to forecast the value of $y_{t+1}$, which is the value of the forecast variable $y$ at time step $t+1$. (c) In an early fusion method, modalities $u$ and $v$ are combined at an early stage of the neural network. (d) In a late fusion method, representations of modalities $u$ and $v$ are combined at one of the final layers of the neural network.}

\label{fig:formulation}
\end{figure}

Here, $u$ and $v$ are fed to different decision functions $f_1$ and $f_2$ respectively. The output of  $f_1$ and $f_2$ are then concatenated and finally passed to the final decision function $f_{late}$, whose output is the forecast value of $\hat{y}_{t+1}$. The formulation is presented in \figref{formulation}.

The error will be determined based on the actual value of a test sample $y_{t+1}$ and the forecast value of the same sample $\hat{y}_{t+1}$. If we report MAE on the whole test set of $m$ examples, 
\begin{equation}
\label{eq:maeeqn}
MAE = \frac{1}{m} \sum_{i=1}^{m} \mid y_{t+1}(i) - \hat{y}_{t+1}(i) \mid
\end{equation}

\subsubsection{Machine Learning Models}
There can be a few different ways to implement our forecasting functions. As the literature on time series forecasting using the type of data that concerns us is limited, a reasonable choice was to use machine learning-based forecasting models. Based on past physical activity and engagement trends, a machine learning model forecasts the physical activity level of the next day. In this case, the forecast variable is the number of steps a user is going to walk the next day. Getting to know this piece of information in advance will give the adaptive intervention agent leverage to prescribe a recommended course of action to keep the user on track to meeting their short-term and long-term goals.

\begin{figure}[tbh!]
     \centering
     \begin{subfigure}{0.14\textwidth}
         \centering
         \includegraphics[width=\textwidth]{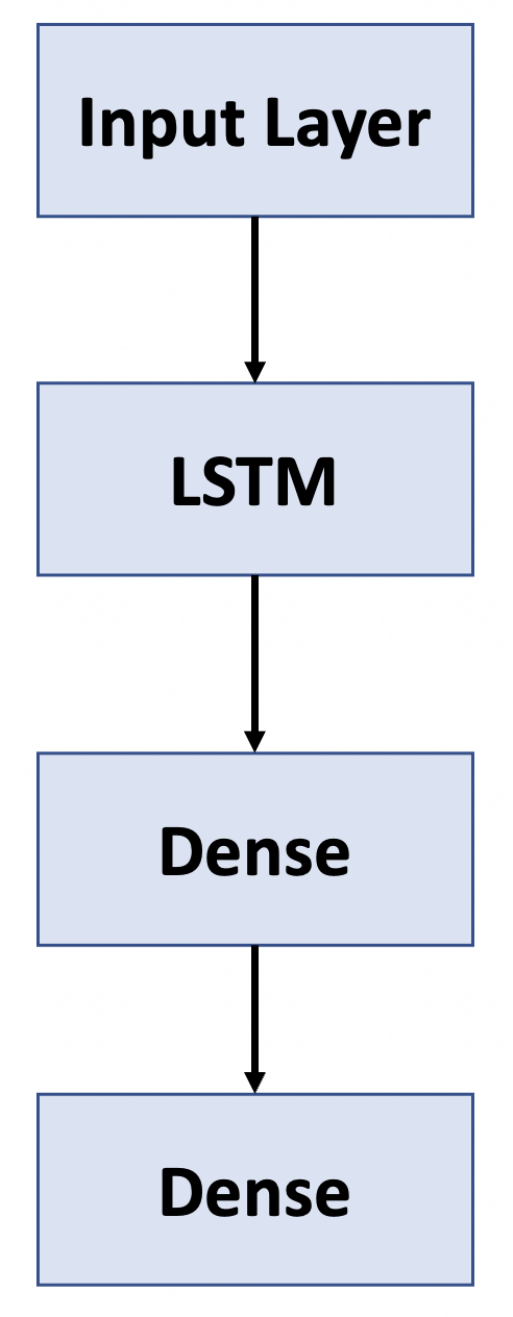}
         \caption{\textit{Early} fusion}
         \label{fig:lstm_early}
     \end{subfigure}
     \hfill
     \begin{subfigure}{0.32\textwidth}
         \centering
         \includegraphics[width=\textwidth]{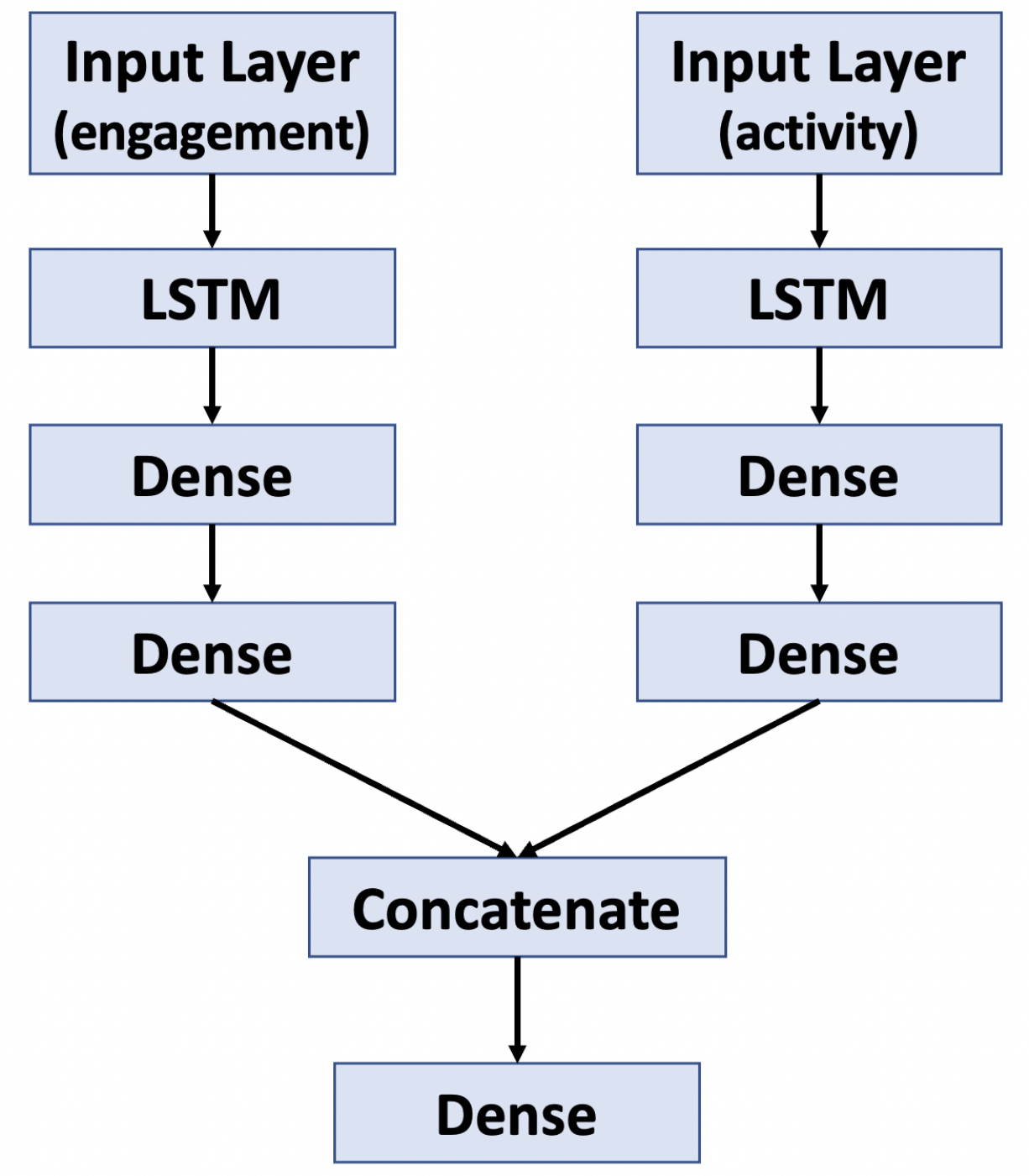}
         \caption{\textit{Late} fusion}
         \label{fig:lstm_late}
     \end{subfigure}
        \caption{Block diagrams of MoveSense's backbones, LSTM-based multimodal forecasting models. (a) Multimodal forecasting with \textit{early} fusion. Here, the input is an early-concatenated feature set that contains the input from both the engagement and physical activity modalities. (b) Multimodal forecasting model with \textit{late} fusion. The neural network processes the inputs from the engagement modality and the activity modality independently until they are concatenated in a later layer.}
        \label{fig:forecasting_models_diagram}
\end{figure}

In the early fusion method, we combine the app engagement features and the physical activity features at the feature level before passing through any neural network layers. If we create similar windows for app engagement and physical activity features, we can easily concatenate them into a single input channel for our machine-learning model. On the other hand, in the late fusion method, we pass each different modality through two independent channels of neural network layers and then pass through temporary decision layers before concatenating the outputs of the two decisions and passing them through a unified decision layer to get the final prediction. As app engagement and physical activity are data from different modalities, it makes sense to operate them separately. The two different multimodal forecasting architectures, one with early fusion and the other with late fusion have been presented in \figref{forecasting_models_diagram}.

\subsubsection{Adaptive Intervention Agent }
The final component of our proposed system is the adaptive intervention design which will be able to provide feedback based on the prediction by the model. Suppose, when a user is not likely to meet the daily goal of their intended level of physical activity according to the forecasting model, the intervention agent will provide them with a reminder so that the user gets the motivation to fulfill their daily goal. The implementation of the intervention agent and the method of choosing the threshold of a forecast being reasonably confident is left for the near future to explore and research.

\subsection{User Studies}
To evaluate our MoveSense system, two user studies were conducted for the experiments that led to this paper. One is BeWell24 \cite{buman2016bewell24} and the other is SleepWell24 \cite{petrov2020rationale}. The BeWell24 study targeted overweight and obese US veterans as they are often at high risk of cardiovascular diseases because of insufficient physical activity and poor quality or duration of sleep \cite{ford2005risks} \cite{nelson2006burden} \cite{johnson2004prevalence}. On the other hand, SleepWell24 was motivated by the fact that sleep apnea is prevalently more common in overweight and obese adults, which can increase the risk of cardiovascular diseases and even premature death of a person, if not treated \cite{vgontzas1994sleep,peppard2000prospective, peker2006increased, yaggi2005obstructive}. The next two sections discuss the studies separately in more detail. The protocol for data collection from human subjects was approved by The Mayo Clinic Institutional Review Board and informed consent was obtained from each participant.

\subsubsection{Prediabetes study}
The prediabetes study, BeWell24, encouraged prediabetic veteran participants to maintain a better lifestyle with the help of an informative smartphone application, hoping to increase the time engaged in light, moderate, and vigorous physical activities. The study was done on two groups of people: the intervention group and the control group. The intervention group used a Fitbit device and a lifestyle intervention app. The control group also used a Fitbit device for recording physical activities but they did not have the lifestyle intervention app. There were 58 people in the intervention group and they volunteered for data collection for over a 9-month duration. The participants were US veterans with prediabetic health conditions.

\subsubsection{Sleep study}
The sleep study, named SleepWell24, recruited participants with obstructive sleep apnea. Similar to the prediabetes study, sleep study participants were also divided into an intervention group and a control group. The intervention group had a lifestyle intervention app and the control group used the Fitbit app. There were 60 people in the intervention group and data were collected over 60 days. The participants were Mayo Clinic patients with obstructive sleep apnea who were prescribed positive airway pressure (PAP) therapy. The SleepWell24 smartphone application was developed to improve adherence to PAP therapy with the help of mHealth technologies \cite{petrov2020rationale}.

\subsection{Data Preprocessing}
For both prediabetes and sleep datasets, we did not have sufficient for several participants. Initially, our prediabetes dataset for the intervention group has engagement and Fitbit data for 58 users and this number is 51 for the sleep dataset. The user studies required that a person would need to wear the Fitbit device for at least 10 hours a day, otherwise, the data for that day would not be considered. It is standard practice to choose 10 hours/day as a threshold for wear time when considering wearable sensor data for further analyses, which is backed by previous studies \cite{troiano2008physical,matthews2008amount, schuna2013adult, sasaki2018number}. Among the 58 prediabetes study users, the average wear time is $12.38 \pm 7.27$ hours per day and the average number of valid days satisfying the 10-hour wear time threshold among users is $155.07 \pm 85.58$ days. On the other hand, the average wear time of the 51 sleep study users is $12.38 \pm 7.28$ hours per day and the average number of valid days for them is $37.84 \pm 20.93$ days. For every user, we eliminated any day where the wear time was less than 10 hours. Furthermore, we set another threshold that a user must have at least 10 days of valid data, otherwise, we eliminate that user. After these two filtering steps, the prediabetes dataset is left with 55 participants, and the sleep dataset is left with 44 participants. A summary of the demographic distribution of the participants of both datasets has been presented in \tblref{demographics}. At first, the app engagement and physical activity data are provided in a minute-by-minute fashion, and we convert it into a daily-level format by aggregating the values. We also create hourly features for minutes used and times opened variables. The summary of the input features is presented in \tblref{feature_summary}. 

\subsection{Experiment Design}
The analysis was completed in multiple steps. At first, the best window size was chosen by training and testing the LSTM models for three feature sets with four different options for the window size: 3, 7, 14, and 21. After training LSTM models independently with four different values for window size as discussed in \secref{window_finder}, we chose the best-found window size, $w$ = 7 for our forecasting. It means that the forecasting was performed by taking the last 7 days’ data as input. The three options for modalities are multimodal early-fusion, only engagement features, and only activity features. Once the best window is found, we train and test ARIMA and linear regression models as baselines and we also develop multimodal late-fusion based LSTM models. From these options, we choose the best model according to its mean absolute error on the test set. Then we train LSTM models for goal-based forecasting that predicts 1-day in advance whether a person's daily number of steps will be above a certain threshold.

\begin{table}[htbp]
\centering
\caption{Statistics on the demographics of the participants after removing participants with insufficient data. Data of 55 participants of the prediabetes dataset and 44 participants of the sleep dataset were used to evaluate our MoveSense system.}
\begin{tabular}{lll}
\hline
Dataset & Prediabetes  & Sleep \\
\hline
N       & $55$      & $44$        \\
Age     & $56.25 \pm 10.49$  & $57.66\pm 13.26$ \\
        & min:$30$, max:$73$ & min:$31$, max:$80$ \\
BMI     & $34.90 \pm 6.92$ & $32.67 \pm 6.55$ \\
        & min:$23.11$, max:$63.71$ & min:$22.82$, max:$52.32$ \\
Gender  & $41$ males & $29$ males \\
        & $14$ females & $15$ females \\
\hline
\end{tabular}
\label{tbl:demographics}
\end{table}

\begin{table}[htbp]
\centering
\caption{All the different engagement and physical activity features have been presented below. Here, Sed, LPA, or MVPA ratio is calculated by dividing the Sedentary, LPA, or MVPA minutes of a particular day by the wear time of that day. For any particular day and user, Sed minutes + LPA minutes + MVPA minutes = wear time. Also, $0 \leq x \leq 1$ for $x \in \{$Sed ratio, LPA ratio, MVPA ratio$\}$. }
\begin{tabular}{ll}
\hline
\textit{Feature type}   & \textit{Dimension} \\
\hline
\textbf{Engagement features}        & \textbf{57} \\
Minutes used                        & 1   \\
Times opened                        & 1   \\
Day of the week (one hot encoded)   & 7   \\
Minutes used in different hours     & 24  \\
Times opened in different hours     & 24   \\
\hline
\textbf{Physical activity features} & \textbf{8} \\
Total steps                         & 1  \\
Sed, LPA, and MVPA minutes           & 3 \\
Total wear time (in minutes)         & 1 \\
Sed, LPA, and MVPA ratios           & 3  \\ 
\hline
\end{tabular}
\label{tbl:feature_summary}
\end{table}

\section{Results and Discussion}
\label{sec:results}
In this section, we present the results and discussion of the experiments to evaluate our MoveSense system. We present the results in the following steps: choosing the best window size, comparing with baselines, the performance of goal-based forecasting, a comparison of unimodal and multimodal systems, different fusion methods of multimodal models, performance for most engaged participants, performance for individual participants, and forecasting additional outcomes.

\subsection{Finding the Best Window Size}
\label{sec:window_finder}
In general, selecting the right window size is an essential step in training a model with time-series data. We train our regression-based forecasting models with different values of window size: 3, 7, 14, and 21. After each training session, the test MAE is calculated. We notice that for both the prediabetes and sleep datasets, window size = 7 had the overall lowest MAE. We present the MAE values for the LSTM model (unimodal and multimodal early fusion) for different window sizes on \tblref{window_chooser}.

\begin{table}[htbp]
\centering
\caption{Mean-absolute-errors for forecasting the next-day number of steps with different window sizes by the LSTM model with multimodal early fusion and unimodal features.}
\begin{tabular}{llllll}
\hline
\textit{Dataset $\downarrow$} & \textit{Window size $\rightarrow$} & \textit{3} & \textit{7}    & \textit{14} & \textit{21} \\ \hline
                 & Multimodal           & 2069       & \textbf{1989} & 2048        & 2004        \\
Prediabetes      & Engagement           & 3199       & 3470          & 3555        & 2745        \\
                 & Activity             & 2084       & 2051          & 2048        & 2130        \\ \hline
                 & Multimodal           & 4595       & 4194          & 5847        & 6048        \\
Sleep            & Engagement           & 4836       & 5354          & 4547        & 5203        \\
                 & Activity             & 4384       & \textbf{4100} & 5646        & 5613   \\  \hline  
\label{tbl:window_chooser}
\end{tabular}
\end{table}

\subsection{Comparison of MoveSense with the Baselines: ARIMA and Linear Regression}
In \tblref{baseline_results} and \tblref{baseline_comparison}, we present a comparative analysis of the LSTM model with linear regression and ARIMA. Our LSTM models outperform both linear regression and ARIMA in terms of mean-absolute-errors (MAE).  The best performance by linear regression was an MAE of 2978 on the prediabetes dataset when using the unimodal activity features. On the other hand, the LSTM performed the best on the prediabetes dataset when both activity and engagement features were used. Early-fusion-based LSTM achieved an MAE of 1989 on the prediabetes dataset, which is 33\% better than the performance of linear regression. On the sleep dataset, a consistent trend was observed with a margin of 13\%. 

\begin{table*}[htbp]
    \centering
    \caption{The multimodal LSTM model of our MoveSense system outperforms all the baselines on the prediabetes dataset and most of the baselines on the sleep dataset in the steps forecasting task. Lin Reg = linear regression model.}
    \begin{tabular}{lllllllll}
    \hline
\textit{}   & \textit{LSTM} & \textit{LSTM} & \textit{LSTM} & \textit{Lin Reg}    & \textit{Lin Reg} & \textit{Lin Reg} & \textit{ARIMA} & \textit{LSTM Late} \\
Dataset $\downarrow$   & \textit{Multimodal}    & \textit{Engagement}    & \textit{Activity}      & \textit{Multimodal} & \textit{Engagement}       & \textit{Activity}         & \textit{Activity}       & \textit{Multimodal}         \\ \hline
Prediabetes & \textbf{1989} & 3470          & 2051          & 2985       & 5390             & 2978             & 3137           & 2813               \\
Sleep       & 4194          & 5354          & \textbf{4100} & 4806       & 6621             & 4809             & 6155           & 4754      \\ \hline        
\end{tabular}
    \label{tbl:baseline_results}
\end{table*}

\begin{table}[h]
\centering
\caption{MAE of MoveSense's multimodal LSTM model with respect to the baselines on the prediabetes and sleep data. $-x$\% means the multimodal LSTM model achieves $x$\% lower MAE than the corresponding baseline. $+y$\% means multimodal LSTM gets $y$\% higher MAE than the baseline.}
\begin{tabular}{lllll}
\hline
\textit{Baseline} $\rightarrow$ & \textit{Unimodal} & \textit{Multimodal} & \textit{ARIMA} & \textit{Multimodal} \\ 
Dataset  $\downarrow$  & \textit{LSTM} & \textit{Lin Reg} & & \textit{LSTM Late} \\
\hline
Prediabetes      & -3\%                   & -33\%     & -37\%               & -29\%                     \\
Sleep            &  +2\%           & -13\%                 & -32\%               & -12\%       \\ \hline     
\end{tabular}
\label{tbl:baseline_comparison}
\end{table}

\subsection{Forecasting for Classification Problems}
The average daily step counts for the prediabetes dataset and sleep dataset participants were 5745 and 7627 respectively. Because of the significant difference between these two values, the classification problems were formulated with different thresholds for the two datasets. Two thresholds, 6000 and 8000, were used for BeWell24 participants and a more standard threshold of 10000 steps was chosen for SleepWell24. The accuracy and F1 score for multimodal (early and late) and unimodal LSTM models are reported in \tblref{classification_summary}. The early-fusion multimodal forecaster gives us an accuracy and f-score of 0.72 and 0.71 respectively on the prediabetes dataset. Also, these metrics are 0.79 and 0.71 respectively for the sleep dataset.

\begin{table}[h]
\centering
\caption{Accuracy and F1 score of MoveSense's LSTM classifiers across different classification thresholds for daily steps goal on the prediabetes and the sleep dataset participants with different modalities. In this table, Multimodal = multimodal early fusion and Late = multimodal late fusion.}

\begin{tabular}{llllll}
\hline
\textit{}   & \textit{} & \multicolumn{2}{c}{\textit{Accuracy}}     & \textit{}     & \textit{} \\
\textit{Dataset}     & \textit{Goal}      & \textit{Multimodal} & \textit{Engagement} & \textit{Activity}      & \textit{Late}      \\ \hline
Prediabetes & 6000      & \textbf{0.72}       & 0.55                & 0.71          & 0.59      \\
Prediabetes & 8000      & 0.71                & 0.68                & \textbf{0.74} & 0.70      \\
Sleep       & 10000     & \textbf{0.79}       & 0.63                & 0.77          & 0.59      \\ \hline
            &           & \multicolumn{2}{c}{\textit{F1 Score}}     &               &           \\
\textit{Dataset}     & \textit{Goal}      & \textit{Multimodal}          & \textit{Engagement}          & \textit{Activity}      & \textit{Late}      \\ \hline
Prediabetes & 6000      & \textbf{0.71}       & 0.19                & \textbf{0.71} & 0.55      \\
Prediabetes & 8000      & \textbf{0.61}       & 0.05                & 0.56          & 0.35      \\
Sleep       & 10000     & \textbf{0.71}       & 0.28                & 0.67          & 0.32    \\ \hline 
\end{tabular}
\label{tbl:classification_summary}
\end{table}

\subsection{Effect of Different Modalities}
\label{sec:res_modality}
In this subsection, we discuss the effects of different modalities on our forecasting problem. We compare the mean-absolute-error (MAE) of prediabetes study participants and sleep dataset participants in \tblref{baseline_results} and \tblref{baseline_comparison}. In \tblref{baseline_results}, we notice that for the prediabetes study participants, the MAEs for multimodal late LSTM, multimodal early LSTM, unimodal (engagement) LSTM, and unimodal (activity) LSTM models are 2813, 1989, 3470, and 2051 respectively. The MAEs for the similar models are 4754, 4194, 5354, and 5354 respectively. For the prediabetes study users, the early fusion-based multimodal model achieved the best error value, while the lowest MAE was achieved by the activity-based unimodal model on the sleep dataset. In \tblref{baseline_comparison}, we notice that the multimodal early LSTM outperforms the best unimodal LSTM by a margin of 3\%, whereas, on the sleep dataset, it achieves comparable performance, with an MAE 2\% higher than the best unimodal model.

We observe the superiority of the multimodal models over unimodal models for linear regression too. In \tblref{baseline_results}, it can be noticed that the multimodal linear regression model outperforms the unimodal counterparts on both prediabetes and sleep datasets.

\subsection{Early Fusion vs Late Fusion}
\label{sec:res_fusion}
One part of our work is to find out which forecasting method, multimodal or unimodal, is superior to the other. But we also need to address which type of fusion method, early or late, we should employ for the multimodal models. We have trained models with early fusion and late fusion.  We have observed consistent results on both datasets. In \tblref{baseline_results} and \tblref{baseline_comparison}, we notice that multimodal early LSTM achieved lower MAE than multimodal late LSTM on both datasets, with margins of 29\% on the prediabetes dataset and 12\% on the sleep dataset.  We observe similar performance on the classification task as well. In \tblref{classification_summary}, we notice that multimodal early fusion achieves higher accuracy and F1 score on both datasets for all the thresholds.

\subsection{Performance Variation for Most Engaged Users.}
\label{sec:most_engaged}
An important question to ask in activity forecasting is does a forecasting model's performance varies significantly for different user groups. To answer the question, the notion of most engaged users has been introduced based on their engagement with the smartphone application. The question still remains on how to choose that subset, i.e., what the threshold should be to decide if a user belongs to that specific group. To answer the question, a series of experiments have been conducted with different subgroups of users: top 75\%, top 50\%, and top 25\% participants based on engagement. From that group, it is seen that on average, the model performs better for the set of top 25\% users, who are above the 75th percentile threshold, in terms of mean absolute error as shown in \tblref{engagement_percentile}. We observe that when we set a threshold of a higher percentile for a participant to be considered, the model's performance seems to improve. It supports our hypothesis that including engagement features can improve the performance of the forecasting problem.

\begin{table}[htbp]
\centering
\caption{Test MAE for steps forecasting based on the selection of most engaged users based on different thresholds of percentiles. The 0th percentile includes all users, and the 25th percentile includes the top 75\% engaged users. Similarly, the 75th percentile includes the top 25\% engaged users.}
\begin{tabular}{llllll}
\hline
\textit{Dataset $\downarrow$} & \textit{Percentile $\rightarrow$} & \textit{0}    & \textit{25}   & \textit{50} & \textit{75} \\ \hline
                 & Multimodal          & 1989 & 1843 & 3522        & \textbf{1677}        \\ 
Prediabetes      & Engagement          & 3470          & 5441          & 3076        & 1917        \\
                 & Activity            & 2051          & 2362          & 2658        & 2638        \\ \hline
                 & Multimodal          & 4194          & 3607          & 4050        & 3763        \\
Sleep            & Engagement          & 5354          & \textbf{2152} & 2571        & 2384        \\
                 & Activity            & 4100          & 4443          & 4232        & 4995      \\ \hline 
\end{tabular}
\label{tbl:engagement_percentile}
\end{table}

\subsection{Forecasting Performance for Individual Participants}
\begin{figure}
    \centering
    \includegraphics[width=0.48\textwidth]{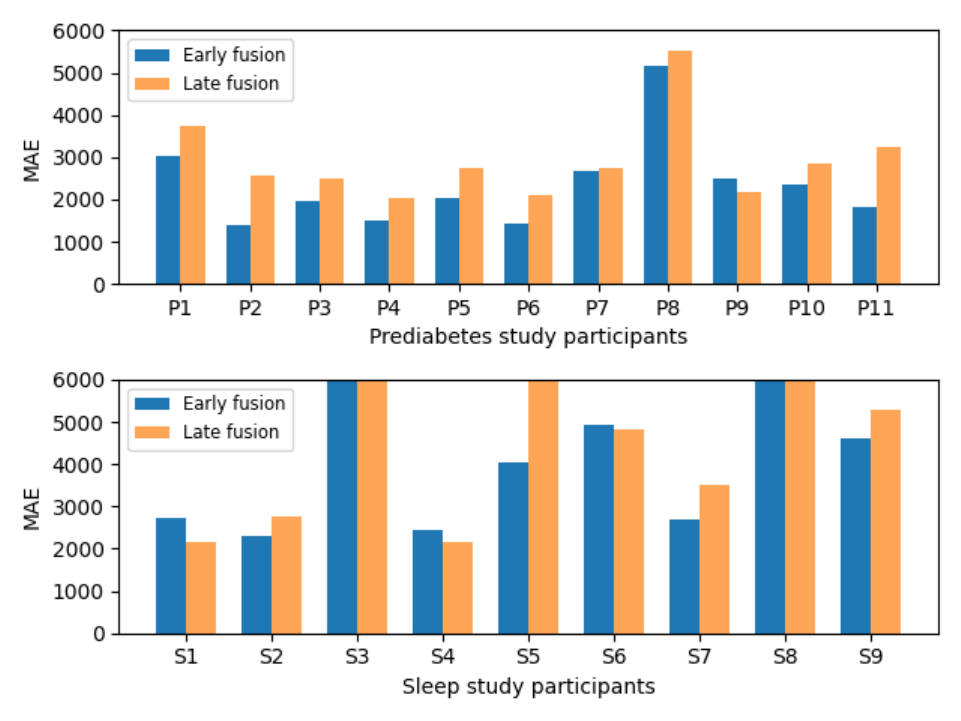}
    \caption{Performances of the MoveSense system's LSTM early fusion and LSTM late fusion multimodal models for steps forecasting on the prediabetes and sleep study participants.}
    \label{fig:mae_per_user}
\end{figure}

From our experiments, we notice that the model's performance varied across different individuals in terms of mean-absolute-errors in the regression setting as shown in \figref{mae_per_user}. For the prediabetes study participants, multimodal early-fusion LSTM gets lower MAE for 10 out of the 11 participants. However, among the sleep study participants, 6 out of 9 participants had lower MAE with early fusion than with late fusion.

Among the prediabetes participants, only 2 out of 11 test set participants had an MAE higher than 3000 with multimodal early fusion. Whereas, for the sleep study participants, 5 out of the 9 participants had an MAE higher than 3000 with multimodal early fusion. We believe that personalization by fine-tuning the models with a specific participant's data can lower the error for that individual \cite{sah2022stressalyzer}.

\subsection{Forecasting Additional Outcomes}

\begin{table}[htbp]
\centering
\caption{Forecasting performance of the MoveSense's multimodal LSTM model (early fusion) on daily sedentary duration, wear time, and duration of light physical activity. NRMSE = normalized root mean squared error.}
\begin{tabular}{llll}
\hline
\textit{Dataset} & \textit{Outcome}   & \textit{MAE} & \textit{NRMSE} \\
\hline
                 & Sedentary duration & 146          & 0.27  \\
Prediabetes      & Wear time          & 136          & 0.18  \\
                 & LPA duration       & 49           & 0.35  \\ \hline
                 & Sedentary duration & 143          & 0.26  \\
Sleep            & Wear time          & 127          & 0.18  \\
                 & LPA duration       & 61           & 0.41 \\ \hline
\end{tabular}
\label{tbl:additional_outcomes}
\end{table}

In this section, we further evaluate our method by forecasting additional outcome variables: daily sedentary duration, light physical activity (LPA) duration, and wear time, i.e., the duration the person is wearing the wearable sensor. We present the results in \tblref{additional_outcomes}. The normalized root mean squared error (NRMSE) was obtained by dividing the root mean squared error (RMSE) by the average value of the corresponding metric. We notice that sedentary duration can be predicted on the day before with an NRMSE of 0.27 and 0.26 on the prediabetes and the sleep datasets respectively. Also, the system is able to forecast LPA duration with MAEs 49 to 61.

\section{Conclusion}
Nonadherence to recommended healthy lifestyle regimens is a persisting healthcare problem and in this busy world, it is very common to be forgetful about completing one's daily need for physical activities. In this work, we have provided a solution model to a multimodal time-series forecasting problem setting and presented experimental results of our proposed model on two important datasets of real subjects. Our datasets were collected in a real-world uncontrolled environment over months, and our models are able to forecast a person's next day's steps with a mean absolute error of 1677 and 2152 for the prediabetes and sleep datasets respectively. We have also provided classification results for different thresholds of daily goals and found that the multimodal forecaster can forecast whether a person will reach their daily goal with an accuracy of 0.72 and 0.79 on the prediabetes and sleep datasets respectively. Finally, our experiments suggest that a window size of 7 is an optimal choice while preparing time-series data for forecasting a person's physical activity. Also, multimodal forecasting models with early fusion are overall a better choice than multimodal forecasting models with late fusion or unimodal forecasting models for forecasting physical activities. We showed how the performance varied for participants above a certain percentile based on app engagement. Finally, we also developed a forecasting model for additional outcomes such as sedentary duration, wear time, and light physical activity duration.

We believe that time-series activity forecasting is an important area of research that can help the development of numerous technologies including adaptive reminders and diet or hydration recommendation systems. Also, other variations of neural network architectures can be implemented and tested to explore if they can improve the performance of forecasting.

\bibliographystyle{./IEEEtran}
\bibliography{./references}

\end{document}